\documentclass{article}

\usepackage{microtype}
\usepackage{graphicx}
\usepackage{subfigure}
\usepackage{booktabs} 

\usepackage{hyperref}
\usepackage{multirow}

\usepackage[accepted]{icml2024}

\usepackage{amsmath}
\usepackage{amssymb}
\usepackage{mathtools}
\usepackage{amsthm}
\usepackage{enumitem}

\usepackage[capitalize,noabbrev]{cleveref}

\theoremstyle{plain}
\newtheorem{theorem}{Theorem}[section]

\theoremstyle{definition}
\newtheorem{definition}[theorem]{Definition}

\theoremstyle{remark}

\usepackage{enumitem}
\usepackage{colortbl}

\usepackage[textsize=tiny]{todonotes}

\icmltitlerunning{Incorporating Retrieval-based Causal Learning with Information Bottlenecks for Interpretable Graph Neural Networks}

\begin{document}

\twocolumn[
\icmltitle{Incorporating Retrieval-based Causal Learning with Information Bottlenecks \\ for Interpretable Graph Neural Networks}



\icmlsetsymbol{equal}{*}

\begin{icmlauthorlist}
\icmlauthor{Jiahua Rao}{sysu}
\icmlauthor{Jiancong Xie}{sysu}
\icmlauthor{Hanjing Lin}{sysu}
\icmlauthor{Shuangjia Zheng}{sjtu}
\icmlauthor{Zhen Wang}{sysu}
\icmlauthor{Yuedong Yang}{sysu,key}
\end{icmlauthorlist}

\icmlaffiliation{sysu}{School of Computer Science and Engineering, Sun Yat-sen University}
\icmlaffiliation{sjtu}{Shanghai Jiao Tong University}
\icmlaffiliation{key}{Key Laboratory of Machine Intelligence and Advanced Computing, Sun Yat-sen University}

\icmlcorrespondingauthor{Yuedong Yang}{yangyd25@mail.sysu.edu.cn}


\vskip 0.3in
]



\printAffiliationsAndNotice{\icmlEqualContribution} 

\begin{abstract}
Graph Neural Networks (GNNs) have gained considerable traction for their capability to effectively process topological data, yet their interpretability remains a critical concern. Current interpretation methods are dominated by post-hoc explanations to provide a transparent and intuitive understanding of GNNs. However, they have limited performance in interpreting complicated subgraphs and can't utilize the explanation to advance GNN predictions. On the other hand, transparent GNN models are proposed to capture critical subgraphs. While such methods could improve GNN predictions, they usually don't perform well on explanations. Thus, it is desired for a new strategy to better couple GNN  explanation and prediction. In this study, we have developed a novel interpretable causal GNN framework that incorporates retrieval-based causal learning with Graph Information Bottleneck (GIB) theory. The framework could semi-parametrically retrieve crucial subgraphs detected by GIB and compress the explanatory subgraphs via a causal module. The framework was demonstrated to consistently outperform state-of-the-art methods, and to achieve 32.71\% higher precision on real-world explanation scenarios with diverse explanation types. More importantly, the learned explanations were shown able to also improve GNN prediction performance.

\end{abstract}

\section{Introduction}
\label{intro}

Graph Neural Networks (GNNs) have made great success in many graph-related tasks, such as node classification~\cite{kipf2017semisupervised}, graph classification~\cite{zhang2018end}, and link prediction~\cite{hamilton2017inductive}, and have been widely used in high-stake applications such as molecular biology~\cite{gilmer2017neural,song2020communicative}, knowledge graphs~\cite{schlichtkrull2018modeling,zheng2021pharmkg}, and social networks~\cite{fan2019graph}. However, their black-box nature has driven the need for explainability, especially in sectors where transparency and accountability are essential like healthcare~\cite{amann2020explainability}, and security~\cite{pei2020amalnet}. Therefore, it is urgent to design interpretable GNN models that can not only achieve superior performances but also interpret the predictions. 



Recently, many efforts have been dedicated to explaining GNNs, including two main categories: \textit{post-hoc} and transparent (\textit{build-in}) methods. 
The \textit{post-hoc} methods identify critical graph structures to best explain  GNN behaviors by training an additional parametric model~\cite{ying2019gnnexplainer, luo2020parameterized, wang2021towards, huang2022graphlime}, or using non-parametric paradigms~\cite{yuan2021explainability,wu2023rethinking}.
Since these methods don't participate in GNN training, they struggle with interpretation accuracy, especially in real-world scenarios with diverse types of explanatory subgraphs. In addition, these methods can't improve GNN performance.
On the other hand, the transparent GNN models integrate the generation of explanations into  GNN training under the proper guidance such as information bottleneck~\cite{wu2020graph,seo2023interpretable} and causality~\cite{lin2021generative, chen2022learning}.
Generally, these methods learn the probability for each edge/node with the attention/masking mechanism and then make predictions via a weighted message-passing scheme with the extracted subgraph. 
Despite their success in improving GNN performance, they inevitably decrease interpretability precision. It is promising to effectively integrate the interpretation and GNN training in a uniform framework.

To this end, we have introduced a novel interpretable GNN framework, denoted RC-GNN, which integrates \textbf{R}etrieval-based \textbf{C}ausal learning into the Graph Information Bottleneck (GIB) theory. 
Specifically, RC-GNN introduces two novel modules: subgraph retrieval and causal graph learning through a semi-parametric paradigm, that 
maximizes the mutual information between the explanatory subgraph and candidate graphs that contain all shared information,
and minimizes the mutual information between the subgraph and the input graph.
Consequently, the retrieval process could capture essential key subgraphs with a comprehensive understanding of the dataset, and the causal framework could optimize the compression of the explanatory subgraphs.


Herein, 
we conducted extensive experiments to evaluate the effectiveness and interpretability of RC-GNN in the graph explanation and classification tasks. RC-GNN was shown consistently to outperform recent state-of-the-art explanation methods, especially in real-world scenarios, and the explanations of RC-GNN also in turn help to improve the classification performance. The improvements were confirmed by visualizations of the extracted subgraph and the learned representations. Overall, RC-GNN was shown to significantly improve the interpretability of GNN, and the performance in classification tasks.

Our contributions are summarized as:
\begin{itemize}
[leftmargin=0in, itemindent=0.1in, topsep=0pt]
    \item We have developed a novel framework for effective GNN explanations and predictions through the incorporation of retrieval-based causal learning into information bottleneck.
    \item  The retrieval process was theoretically proved able to combine with causal learning for GNN explainability. 
    \item The proposed framework was indicated to outperform SOTA methods through extensive experiments on explanation and classification tasks, and have an average precision increase of 32.71\% on the real-world explanation scenarios.

\end{itemize}

\section{Related Works}

\subsection{Explainability of GNNs}
The explainability of GNNs has received substantial attention in machine learning in recent years.  
Most of existing works~\cite{ying2019gnnexplainer, yuan2020xgnn, luo2020parameterized, vu2020pgm, wang2021towards, lucic2022cf, huang2022graphlime} focus on training an additional parameterized model to identify crucial nodes, edges, or subgraphs that explain the behavior of target GNNs for input samples.
For example, GNNExplainer~\cite{ying2019gnnexplainer} learns graph/feature masks to select important subgraphs and mask out of less important nodes. 
ReFine~\cite{wang2021towards} combines contrastive learning into class-wise generative probabilistic models to generate multi-grained explanations.
GraphLIME~\cite{huang2022graphlime} learns a nonlinear interpretable model in the subgraph of the node being explained. 
Although many successes, their produced explanations are heavily dependent on hyperparameter choices of secondary models. 
Moreover, these explainers’ black-box nature raises doubts about their ability to provide comprehensive explanations for GNN models.

Instead of creating parametric explanation models, the other lines are non-parametric explanation methods that do not involve any additional trainable models. For example, 
SA~\cite{baldassarre2019explainability} and GradCAM~\cite{pope2019explainability} employ heuristics like gradient propagation as feature contributions of specific instances. 
SubgraphX~\cite{yuan2021explainability} generates subgraph explanations with the help of the Monte Carlo Tree Search, which unavoidably results in a high computational cost. 
MatchExplainer~\cite{wu2023rethinking} employs a subgraph-matching paradigm to explore commonly explanatory subgraphs, but is mostly successful in synthetic datasets with clear and simple subgraph patterns.
These reinforce a need for our proposed retrieval-based causal method that could perform consistently well on both simple and complex datasets.


\subsection{Interpretable Graph Learning}
Interpretable graph neural networks have recently attracted growing attention in graph representation learning,  aiming to integrate the generation of explanations into the model training process.
Such approaches~\cite{gat2018graph, xie2018crystal}  augment interpretability via attention mechanisms, but they are either limited to specific GNN architectures or cannot indicate important graph structures for predictions.

To circumvent the issues,  it turns popular to select a neighborhood subgraph of a central node for message passing under the proper guidance such as information bottleneck~\cite{wu2020graph, yu2021graph} and causality~\cite{sui2022causal, fan2022debiasing}. GIB~\cite{yu2021graph} recognizes the maximally informative IB-subgraph for the improvement of graph classification and graph interpretation. CAL~\cite{sui2022causal} incorporates the attention learning strategy with causal theory on graphs to achieve stable performances of GNN. 
PGIB~\cite{seo2023interpretable} involves the information bottleneck framework to demonstrate the prototypes with the key subgraph from the input graph that is important for the model prediction.
Unfortunately, they inevitably decrease interpretability precision due to the lack of an effective framework to integrate graph explanations and predictions.

Here, we propose to retrieve informative subgraphs and compress the informative subgraph with the causal module by the backdoor adjustment of causal theory. 
Such settings ensure a semi-parametric training paradigm to integrate subgraph retrieval with causal learning in the GIB theory for graph classification and explanation.


\section{Preliminaries}
In the context of RC-GNN, we involve retrieval-based causal learning with the GIB theory to extract meaningful explanations and in turn help to improve GNN performances. 
In the following section, we will introduce notations used throughout the paper, followed by the definitions of the GIB theory and the causal view on GNNs.

\subsection{Explanations for GNNs}
Let $\mathcal{G}$ denote a graph on nodes $\mathcal{V} = \{v_1, v_2, ..., v_N \} $ and edges $\mathcal{E} \subset \mathcal{V} \times \mathcal{V} $ that are associated with \textit{d}-dimensional node features $H \in \mathbb{R}^{N \times d}$. Without loss of generality, we consider the explaining task of GNNs to generate a faithful explanation for each $\mathcal{G} = ( \mathcal{V}, \mathcal{E} )$ by identifying a subset of nodes and edges $\mathcal{G}_{S}$, which are essential to the predictions.

The GNN model is used to learn a function $f(\mathcal{G})$, which maps the input graph $\mathcal{G}$ to a target output $h_i^{(l)}$ : 
\begin{equation}
{
h^{(l)}_{i} = {\rm Comb}^{(l)} {\left( h_i^{(l-1)}, {\rm Aggr}^{(l)} {\left( h_j^{l-1}, \forall v_j \in \mathcal{N}_i \right)} \right)}
}
\end{equation}
where $h^{(l)}_i$ are the representation of node $v_i$ at the $l$-th layer, $\mathcal{N}_i$ is the neighbourhood set of node $v_i$, and $h^{(0)}_i$ is initialised with its node properties. ${\rm Aggr}$ and ${\rm Comb}$ stand for the aggregation  and   update functions, respectively. 

After $l$ graph convolutions, $h^{(l)}$ have captured their $l$-hop neighborhood information. Finally, a readout function is used to aggregate all node representations output by the $l$-th GNN layer to obtain the entire graph representation ${ H}$:
\begin{equation}
   { H} = \sum_{v_i \in \mathcal{V}} {\rm Readout}(h^{(l)}_i).
\end{equation}

\subsection{Graph Information Bottleneck}

\begin{definition}
(Graph Information Bottleneck).~\cite{yu2021graph} Given a graph $\mathcal{G}$ and its label ${Y}$, the  Graph Information Bottleneck (GIB) principle seeks for the most informative yet compressed subgraph $\mathcal{G}_{S}$ by optimizing the following objective:
\begin{equation}
\label{eq:gib}
    \mathop{\rm max} \limits_{\mathcal{G}_{S} \subset \mathcal{G} } \left[  I (Y; \mathcal{G}_{S}) - \beta I(\mathcal{G}; \mathcal{G}_{S}) \right]
\end{equation}
where $\mathcal{G}_{S}$ indicates the key subgraphs and $Y$ is graph labels.
\end{definition}

The first term maximizes the mutual information between the graph label and the compressed subgraph,  ensuring that the compressed subgraph contains maximal information for predicting the graph label. The second term minimizes the mutual information between the input graph and the compressed subgraph,  ensuring that the compressed subgraph contains minimal information about the input graph.

\subsection{Causal View on GNNs}
Following previous studies~\cite{sui2022causal, fan2022debiasing}, we take a causal look at the GNN modeling to construct a Structural Causal Model (SCM). The model presents the causalities among five variables: graph data, causal variable, shortcut variable, graph representation, and prediction, where the link from one variable to another indicates the cause-effect relationship. We list the following explanations for SCM:
\begin{itemize}[leftmargin=0in, itemindent=0.1in, topsep=0pt,parsep=0pt]
    \item \textbf{C $\leftarrow$ G $\rightarrow$ S.} The causal variable $C$ that truly reflects the intrinsic property of the graph data $G$.  $S$ represents the shortcut feature that is usually caused by data biases or trivial patterns.
    \item \textbf{C $\rightarrow$ H $\leftarrow$ S.} The variable $H$ is the representation of the given graph data $\mathcal{G}$. To generate $H$, the GNN model $f$ takes the shortcut feature $S$ and the causal feature $C$ as input to distill discriminative information.
    \item \textbf{H $\rightarrow$ Y.} The ultimate goal of graph representation learning is to predict the label of the input graphs. The classifier will make prediction ${Y}$ based on the graph representation $H$.
\end{itemize}

\begin{figure*}[t]
\begin{center}
\centerline{\includegraphics[width=\linewidth]{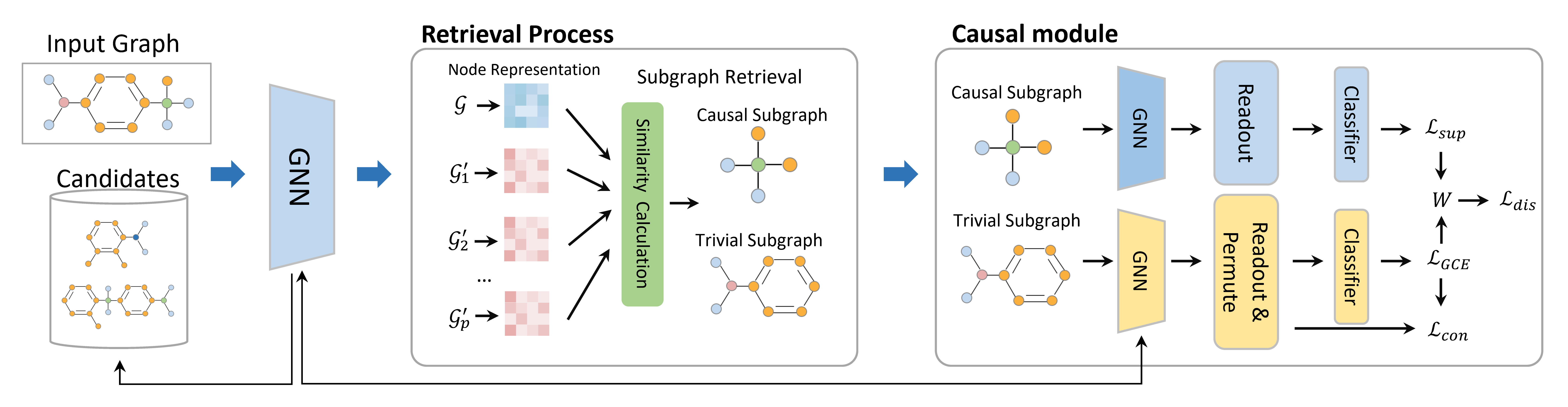}}
\caption{The overall framework of our method.}
\label{fig1}
\end{center}
\vskip -0.2in
\end{figure*}

\subsection{Theoretical Analysis}
\label{sec-theo}

Our method is a novel explainable GNN framework that integrates the retrieval-based causal learning within the GIB theory, thereby enabling the retrieval process to capture the essential causal subgraph detected by the GIB theory. More precisely, we reformulate the GIB objective shown in Eq.~\ref{eq:gib} as it is hard to optimize due to the discrete nature of irregular graph data.

\paragraph{Maximizing $I (Y, \mathcal{G}_{S})$}
We firstly examine the objective of $I (Y, \mathcal{G}_{S})$.
This term measures the relevance between $\mathcal{G}_{S}$ and $Y$. Instead of merely optimizing the information hidden in $\mathcal{G}_{S}$, we reformulated this term into $I(Y; \mathcal{G}_{S}, \mathcal{G}^{'} ) + I(\mathcal{G}^{'}, \mathcal{G}_{S})$ under the hypothesis that graphs of the same class typically share common motif patterns. To be specific, they aim to explore the explanation of $\mathcal{G}$ from the candidate set $\mathcal{D} = \{\mathcal{G}^{'}, f({\mathcal{G}^{'}}) = Y \}$ with the same predicted labels $\hat{Y}$ (refer to Appendix~\ref{sup-meth} for detailed proofs): 

\begin{equation}
\begin{aligned}
\label{eq:first}
    \mathop{\rm max} \limits_{\mathcal{G}_{S} \subset \mathcal{G} } I (Y, \mathcal{G}_{S}) &= \mathop{\rm max} \limits_{\mathcal{G}_{S} \subset \mathcal{G} } \left[ I(Y; \mathcal{G}_{S}, \mathcal{G}^{'} ) + I(\mathcal{G}^{'}, \mathcal{G}_{S}) \right] \\
    &:= \mathop{\rm max}\limits_{\mathcal{G}^{'} \subset \mathcal{D} } \left[ 
        \mathop{\rm max} \limits_{\mathcal{G}_{S} \subset \mathcal{G}, \mathcal{G}^{'}_{S} \subset \mathcal{G}^{'} } I (\mathcal{G}^{'}_{S}; \mathcal{G}_{S})
    \right]
\end{aligned}
\end{equation}

Similar to the information bottleneck theory~\cite{achille2018emergence}, we define the sufficient explanations as:

\begin{definition}
(Sufficient Explanation)~\cite{wu2023rethinking}. Given a graph $\mathcal{G}^{'}$ from the candidate set $\mathcal{D}$, the explanation ${\mathcal{G}}^{suf}_{S}$ of $\mathcal{G}$ is sufficient if and only if $I(\mathcal{G}^{'}, {\mathcal{G}}^{suf}_{S}) =  I( \mathcal{G}^{'},  \mathcal{G}) $.
\end{definition} 

The sufficient explanation ${\mathcal{G}}^{suf}_{S}$
of $\mathcal{G}$ keeps all joint information with $\mathcal{G}^{'}$ related to the GNN model $f$. In other words, ${\mathcal{G}}^{suf}_{S}$ contains all the shared information between $\mathcal{G}$ and $\mathcal{G}^{'}$. 
Symmetrically, the sufficient explanation for $\mathcal{G}^{'}$ satisfies $I(\mathcal{G}^{'}, {\mathcal{G}}_{S}^{'suf}) =  I( \mathcal{G}^{'},  \mathcal{G}) $.
Therefore, we utilize the retrieval subgraph process across the candidate set, which is committed to maximizing Eq.~\ref{eq:first}, to generate sufficient explanations for the input graph.

\paragraph{Minimizing $I(\mathcal{G}, \mathcal{G}_{S})$}
For the second term in Eq.~\ref{eq:gib}, the goal is to extract an informative subgraph $\mathcal{G}_{S}$ from $\mathcal{G}$ that contains minimal information about $\mathcal{G}$. 
From the causal view on GNNs, we have realized that shielding the GNNs from the confounder $\mathcal{G}_t = \mathcal{G} - \mathcal{G}_{c}$ is the key to exploiting causal subgraph $\mathcal{G}_{c}$. 
We should achieve the graph representation by eliminating the backdoor path with causal theory, which provides us with a feasible solution. We can exploit the do-calculus on the variable to remove the backdoor path by estimating~\cite{sui2022causal}:
\begin{equation}
\label{eq:backdoor}
\begin{aligned}
    P(\mathcal{G} \vert  do(\mathcal{G}_{S}) ) &= P({\mathcal{G} \vert  \mathcal{G}_{S}}) \\
        & = \sum\limits_{\mathcal{G}_t \subset \mathcal{D}_{t}} { P(\mathcal{G} \vert  \mathcal{G}_{S}, \mathcal{G}_t) P(\mathcal{G}_t \vert \mathcal{G}_{S})}  \\ 
        &= \sum\limits_{\mathcal{G}_t \subset \mathcal{D}_{t}} { P(\mathcal{G} \vert  \mathcal{G}_{S}, \mathcal{G}_t) P(\mathcal{G}_t)} \\  %
        &= \sum\limits_{\mathcal{G}_t \subset \mathcal{D}_{t}} { P(\mathcal{G} \vert  \mathcal{G}_{S}, \mathcal{G}_t) P(\mathcal{G}_t)}
\end{aligned}
\end{equation}
where $\mathcal{D}_{t}$ denotes the confounder set and $P(\mathcal{G} \vert  \mathcal{G}_{S}, \mathcal{G}_t)$ represents the conditional probability given the causal subgraph $\mathcal{G}_{S}$ and confounder $\mathcal{G}_t$; $P(\mathcal{G}_t)$ is the prior probability of the confounder. Eq.~\ref{eq:backdoor} is called the backdoor adjustment.

Recent studies on contrastive learning have proven that minimizing contrastive loss is equivalent to maximizing the mutual information between two variables.  
Hence, we incorporate the contrastive learning between the causal subgraph and trivial subgraph for implementing Eq.~\ref{eq:backdoor}.

%




\section{Methodology}

In this section, we first briefly introduce the model architecture of our framework in Figure~\ref{fig1}. Then we describe the proposed subgraph retrieval module, followed by the causal graph learning for our causal assumption. 

\subsection{Model Architecture}

Figure~\ref{fig1} presents an overview of our method. We first generate the representations of nodes in the input graph $\mathcal{G}$ using a GNN encoder. The candidate set $\mathcal{D}$ also is generated from the GNN encoder that shares a common category.
Then, the node representations are passed to the subgraph retrieval layer that finds sufficient subgraph explanations from the candidate set. Next, we apply the causal theory to minimize the mutual information between the input graph and the key subgraph, ensuring the compressed subgraph contains minimal information about the input graph. Finally, we export the subgraph as the explanations for the input graph and also merge them into the causal module for the final predictions.

\subsection{Subgraph Retrieval}
At the core of our method is the idea of incorporating a non-parametric subgraph retrieval with parametric causal graph learning to explore the most informatively joint substructure for the input instances.
As analyzed before, the first goal is to find a subgraph $\mathcal{G}_{S}$ with $K$ nodes to maximize $I (\mathcal{G}^{'}_{S}, \mathcal{G}_{S})$. Due to the hypothesis that the optimal counterpart $\mathcal{G}^{'}$ ought to share the same explanatory substructure as $\mathcal{G}$. Our target is equivalent to optimize $I (\mathcal{G}^{'}_{S}, \mathcal{G}_{S})$ with $\mathcal{G}_{S} \subset \mathcal{G}$ and $\mathcal{G}_{S}^{'} \subset \mathcal{G}^{'}$.
Instead of directly matching subgraphs from the $\mathcal{G}^{'}$ with node correspondence-based distance for measuring, we applied the retrieval strategy by explicitly asking the model to decide which key substructures to retrieve.

To be specific, given the input graph $\mathcal{G}$ and the $\mathcal{G}^{'}$ from candidate set $\mathcal{D}$, $s_{\mathcal{G}}$ is defined and minimized as follows:

\begin{equation}
     {s_{\mathcal{G}} (\mathcal{G} , \mathcal{G}^{'} ) } = \mathop{\rm max}\limits_{\mathcal{G}_{c}, \mathcal{G}^{'}_{c}} \sum_{i \in \mathcal{G}_c, j \in \mathcal{G}^{'}_c } {\rm sim} ( \mathcal{G}_{c_i}, \mathcal{G}^{'}_{c_j} )
\end{equation}

where $h_{\mathcal{G}_{c}}$ is encoded from the learned GNN model $f$,  $s_{\mathcal{G}} $ is the vector similarity measured by Euclidean distance.

After the pairwise subgraph similarity retrieval, we calculate the corresponding node score in $\mathcal{G}$ across the entire candidate set. Notably, not all graphs in $\mathcal{D}$ are qualified counterparts unless the normalized similarity score $_{\mathcal{G}} (\mathcal{G} , \mathcal{G}^{'} ) $ above the threshold $t$ (i.e. $t$=0.4). Besides, the $\mathcal{G}^{'}$ needs to have correct graph label prediction with the current GNN model.

Therefore, our method can provide many-to-one explanations for a single graph $\mathcal{G}$ once a bunch of candidate subgraphs is given. This offers a new understanding that the determinants for GNNs’ predictions are diverse, and GNNs can gain correct predictions based on several different explanatory subgraphs.




\subsection{Causal Graph Learning}

Motivated by the above causal analysis, in this section, we present our proposed causal module, to ensure the compressed subgraph contains minimal information. Herein, we utilize two separate GNN modules with their corresponding subgraphs (causal and trivial graph) to encode the corresponding substructure into disentangled representations, respectively. Then we permute the trivial representations among the training graphs with contrastive learning so that the correlation between graph representations and trivial representations is removed.

\paragraph{Learning Disentangled Graph Representations} Until now, we have retrieved the sufficient causal graph $\mathcal{G}_{S}$ and the trivial graph $\mathcal{G}_{t} = \mathcal{G} - \mathcal{G}_{S}$. Now we need to make the causal and trivial graphs to capture the causal and shortcut features from the input graphs, respectively.
Inspired by \citet{fan2022debiasing}, our method simultaneously trains two separate GNNs $f_{c}, f_{t}$ with linear classifiers $C_{c}, C_{t}$:
\begin{equation}
    h_{\mathcal{G}_c} =  {f_{c}} \left( h_{\mathcal{G}}, \mathcal{G}_c\right),  \hat{Y}_{\mathcal{G}_c} = C_{c} (h_{\mathcal{G}_c})
\end{equation}
\begin{equation}
    h_{\mathcal{G}_t} =  {f_{t}} \left( h_{\mathcal{G}}, \mathcal{G}_t \right), \hat{Y}_{\mathcal{G}_t} = C_{t} (h_{\mathcal{G}_t})
\end{equation}

Therefore, GNN $f_{c}$ and $f_{t}$ embed the corresponding subgraphs into causal embedding $h_{\mathcal{G}_c}$ and trivial embedding $h_{\mathcal{G}_t}$, respectively. Subsequently, concatenated vector $h = [h_{\mathcal{G}_c}, h_{\mathcal{G}_t}]$ is fed into linear classifier to predict the target label $Y$.
Following the previous study~\cite{fan2022debiasing}, we first utilize the generalized cross entropy (GCE) loss to amplify the trivial information of the trivial GNN:
\begin{equation}
    \mathcal{L}_{GCE} ( C_{t} (h_{\mathcal{G}_t}), Y ) = \frac{1}{q} { (1 - C_{t} (h_{\mathcal{G}_t})^{q} ) }
\end{equation}

The causal graph aims to estimate the causal features, so we classify its representation to the ground-truth label with $\mathcal{L}_{sup} = {\rm CE} (h_{\mathcal{G}_c}, Y ) $. Thus, we also train a causal GNN simultaneously with the weighted cross-entropy loss. The graphs with high probability from the trivial classifier $C_{t}$ can be regarded as unbiased samples. In this regard, we could obtain the disentangled score of each graph as:
\begin{equation}
    W(h) = \frac{ {\rm CE} (\hat{Y}_{\mathcal{G}_c}, Y) }{ {\rm CE} (\hat{Y}_{\mathcal{G}_t}, Y) + {\rm CE}(\hat{Y}_{\mathcal{G}_c}, Y) }
\end{equation}
\begin{equation}
    \mathcal{L}_{dis} = W(h) {\rm CE} (\hat{Y}_{\mathcal{G}_c}, Y)  + \mathcal{L}_{GCE} (\hat{Y}_{\mathcal{G}_c}, Y)
\end{equation}
where ${\rm CE}$ represents the calculation of cross-entropy loss.

\paragraph{Counterfactual Contrastive Learning}
As discussed in Section~\ref{sec-theo}, one promising solution to alleviating the confounding effect is the backdoor adjustment — that is minimizing the contrastive loss between the causal subgraph and trivial subgraph.

More specifically, we randomly permute trivial embeddings in each mini-batch and obtain $h_{p} = [h_c, \hat{h}_t]$, where $\hat{h}_t$ represents the randomly permuted trivial vectors of $h_t$. As $h_c$ and $\hat{h}_t$ in $h_{p}$ are randomly combined from different graphs, they will have much less correlation than $h = [h_c; h_t]$ (negative pairs) where both are from the same graph. To make $f_{t}$ and $C_{t}$ still focus on the trivial information, we also swap label $Y$ as $Y^{'}$ along with $h_t$, so that the spurious correlation between $h_t$ and $Y$ still exists. With the generated unbiased samples, we utilize the following loss function: 
\begin{equation}
\label{eq:con_l}
    \mathcal{L}_{con} = W(h) {\rm CE}(C_{c}(h_{p}), Y) + \mathcal{L}_{GCE}(C_{t}(h_{p}), Y^{'})
\end{equation}

We define the Eq.~\ref{eq:con_l} as the causal contrastive loss. It pushes the predictions of such intervened graphs to be invariant and stable across different stratifications, due to the shared causal features.

Together with the supervised loss and disentanglement loss, the total loss function is defined as:
\begin{equation}
\label{eq-total}
    \mathcal{L} = \mathcal{L}_{sup} + \lambda_1 \mathcal{L}_{dis} + \lambda_2 \mathcal{L}_{con}
\end{equation}
where $\lambda_1 $ and $\lambda_2$ are hyper-parameters that determine the importance of disentanglement and causal contrastive learning, respectively.

\section{Experiments}

\begin{table*}[t]
\caption{Explainability performance for our method and other baseline explainers on various benchmark}
\label{xai-table}
\vskip -0.08in
\begin{center}
\resizebox{\textwidth}{!}{
\begin{tabular}{lccccccccccccccc}
\toprule
\multirow{3}{*}{Benchmark} & \multicolumn{3}{c}{BA3-Motif} & \multicolumn{3}{c}{MUTAG} & \multicolumn{3}{c}{Benzene} & \multicolumn{3}{c}{MutagenicityV2} \\
 & \multicolumn{3}{c}{Barabasi-Albert graph}  & \multicolumn{3}{c}{${\rm NH_2}$, ${\rm NO_2}$ chemical group} & \multicolumn{3}{c}{Benzene Ring} &  \multicolumn{3}{c}{46 types of ToxAlerts} \\ 
 & ACC-AUC & Rec@5 & Prec@5 & ACC-AUC & Rec@5 & Prec@5 & ACC-AUC & Rec@5 & Prec@5 & ACC-AUC & Rec@5 & Prec@5  \\
\midrule
GradCAM  & 0.533& 0.212& 0.385 & 0.786 & 0.129& 0.120 & {0.939}& {0.310}& {0.943}& 0.681& 0.158& 0.242 \\
SAExplainer  & 0.518& 0.243& 0.512 & 0.769& 0.329& 0.240 & 0.932& 0.307& 0.946 &0.685& 0.258& 0.345 \\
GNNExplainer  & 0.528& 0.157& 0.576 & 0.895& 0.291& 0.141 & 0.943& 0.320& 0.866&\textbf{0.845}& 0.228& 0.315\\
PGExplainer & 0.586& 0.293& 0.769 & 0.631& 0.029& 0.160 & 0.946& 0.299& 0.864& 0.668& 0.290& 0.283 \\
PGMExplainer & 0.575 & 0.250& 0.835 & 0.714& 0.150& \underline{0.242} & 0.751& 0.119& 0.373& 0.712& 0.173& 0.239 \\
ReFine  & 0.576& 0.297& \underline{0.872} & 0.955& 0.454& 0.160 & \underline{0.955}& \underline{0.335}& \underline{0.974}& 0.786& \underline{0.310}& \underline{0.383} \\
MatchExplainer & \underline{0.639}& \underline{0.305}& {0.834} & \textbf{0.997}& \underline{0.424}& {0.234} & 0.932& 0.314& {0.966} & 0.779& 0.284& {0.303}  \\
\midrule
RC-Explainer & \textbf{0.654} & \textbf{0.306} \footnote{} & \textbf{0.883} &  \underline{0.992}& \textbf{0.475} & \textbf{0.282} & \textbf{0.984} & \textbf{0.353}& \textbf{0.997} &\underline{0.826} & \textbf{0.335} & \textbf{0.518} \\
Relative Impro. (\%) & 2.347 & 0.328& 1.261 & -& 4.623 & 16.53 & 3.037 & 5.373 & 2.361&-& 8.064& 35.24 \\
\bottomrule
\end{tabular}
}
\end{center}
\vskip -0.2in
\end{table*}

\subsection{Datasets and Experimental Settings}

\subsubsection{Datasets}
We consider two explanation benchmarks with GNNs and also benchmarking two new real-world datasets for explaining, following  \citet{agarwal2023evaluating} and \citet{rao2022quantitative}:
    
\textbf{Motif graph classification.} \citet{wang2021towards} create a synthetic dataset, BA-3Motif, with 3000 graphs. They use the Barabasi-Albert (BA) graphs as the base and attach each base with one of three motifs: house, cycle, or grid. We use the ASAP model~\cite{ranjan2020asap} that realizes a 99.75\% testing accuracy for baseline methods.

\textbf{Molecular benzene graph classification.} The Benzene~\cite{sanchez2020evaluating, rao2022quantitative} dataset contains 12,000 molecular graphs extracted from the ZINC15~\cite{sterling2015zinc} database and labeled into two classes where the task is to identify whether a given molecule has a benzene ring or not. The ground truth explanations are the nodes comprising the benzene rings.

\textbf{Molecular mutagenic graph classification.} The MUTAG dataset contains 4,337 graph molecules labeled into two different classes according to their mutagenic properties. The carbon rings with both ${\rm NH_2}$ or ${\rm NO_2}$ chemical groups are valid explanations for the GNN model~\cite{xu2018how} to recognize a given molecule as mutagenic~\cite{agarwal2023evaluating}. Furthermore, \citet{rao2022quantitative} created a newly ames-mutagenic molecular dataset, namely MutagenicityV2, including 6,506 compounds and corresponding Ames mutagenicity values. And \textbf{46} toxic substructures were summarized by \citet{sushko2012toxalerts}. These handcrafted substructures would be used as ground truths for explanations.

\textbf{Molecular activity graph classification.}
We also compiled another more challenging dataset to automatically determine substructures without handcrafted rules. For this purpose, we employed the property cliffs in the Cytochrome P450 3A4 (CYP3A4) inhibitions experimentally measured by \citet{veith2009comprehensive} which includes 3,626 active inhibitors/substrates and 5,496 inactive compounds. The active compounds were compared with inactive ones through MMPA~\cite{hussain2010computationally} in the RDKit package. This led to 106 molecular pairs and corresponding substructures.

\subsubsection{Baselines}

We compared our method with the state-of-the-art baselines:






\textbf{Explanation.} {SA}~\cite{baldassarre2019explainability} and{Grad-CAM}~\cite{pope2019explainability} use the gradients to produce the feature contribution for explanations.
{GNNExplainer}~\cite{ying2019gnnexplainer}, {PGExplainer}~\cite{luo2020parameterized}, and {PGM-Explainer}~\cite{vu2020pgm} train additional parameterized models to identify crucial subgraphs that explain the behavior of target GNNs.
{ReFine}~\cite{wang2021towards} exploits the pre-training and fine-tuning idea to develop a multi-grained GNN explainer. 
{MatchExplainer}~\cite{wu2023rethinking} proposed a non-parametric subgraph matching framework to explore explanatory subgraphs.

\textbf{Interpretable GNNs.} {GIB}~\cite{yu2021graph} proposed a GIB framework to discover an IB-subgraph for the improvement of graph classification. {CAL}~\cite{sui2022causal} introduced a casual attention learning strategy to achieve stable performance of GNNs. {DisC}~\cite{fan2022debiasing} proposed a general disentangled GNN framework to learn the causal substructure and bias substructure, respectively, for superior generalization performance.

\begin{table*}[t]
\caption{Graph Classification performance for our method and other baselines on various benchmark}
\vskip 0.05in
\label{gnn-table}
\begin{center}
\resizebox{0.95\linewidth}{!}
{
\begin{tabular}{lcc|cccc|ccc|c}
\toprule
\multirow{2}{*}{Benchmark} & \multirow{2}{*}{Metric} & \multirow{2}{*}{Backbone} & \multicolumn{4}{c}{Augmentation with Explainers} & \multicolumn{3}{|c|}{Interpretable GNNs} & \\

 & & & Original & DropNode & PGDrop & MatchDrop & GIB & CAL & DisC & RC-GNN \\

\midrule

\multirow{2}{*}{MUTAG} & \multirow{2}{*}{ACC} & GCN & 0.828 & 0.832 & 0.825 & {0.844} & 0.776 & 0.892 & 0.873 &  \textbf{0.930 $\pm$ 0.009 }\\
& & GIN & 0.832 & 0.835 & 0.838 & {0.838} & 0.839 & 0.899 & 0.882 &  \textbf{0.932 $\pm$ 0.007} \\

\midrule

\multirow{2}{*}{MutagenicityV2} & \multirow{2}{*}{ACC} & GCN & 0.802  & 0.776 & 0.756 & 0.805 & 0.805 & 0.885 & 0.854 & \textbf{0.936 $\pm$ 0.004}\\
& & GIN & 0.817  & 0.801 & 0.763 & 0.813 & 0.823 & 0.902 & 0.873 & \textbf{0.943 $\pm$ 0.006}\\

\midrule



\multirow{2}{*}{HIV} & \multirow{2}{*}{AUC} & GCN & 0.764 & 0.694 & 0.711 & 0.744 & 0.767 & 0.783 & 0.773 & \textbf{0.814 $\pm$ 0.003}   \\
& & GIN & 0.770 & 0.700 & 0.708 & 0.750 & 0.784 & 0.804 & 0.794 & \textbf{0.820 $\pm$ 0.002}\\ 

\midrule

\multirow{2}{*}{BBBP} & \multirow{2}{*}{AUC} & GCN & 0.672 & 0.644 & 0.658 & 0.704  & 0.664 & 0.723 & 0.711 & \textbf{0.748 $\pm$ 0.006}  \\
& & GIN & 0.678 & 0.650 & 0.662 & 0.719 & 0.677 & 0.720 & 0.714 & \textbf{0.756 $\pm$ 0.006}  \\

\bottomrule
\end{tabular}
}
\end{center}
\vskip -0.1in
\end{table*}

Further details of the datasets, baselines, and implementation details are presented in the Appendix~\ref{sup-impl}.

\subsubsection{Evaluation Metrics}
In this paper, we used three widely used evaluation metrics to quantitatively evaluate the quality of explanations:

\textbf{Predictive Accuracy (ACC@$\rho$)} It measures the fidelity of the explanatory subgraphs by feeding it solely into the target model and auditing how well it recovers the target prediction. We report the average ACC@$\rho$ over all graphs in the testing sets, and further denote ACC-AUC as the area under the ACC curve over different selection ratios $ \in \{0.1, 0.2, ..., 0.9, 1.0\}$. ACC@$\rho$ and ACC-AUC are suitable for all the datasets.
As suggested in prior studies~\cite{wang2021towards, wu2023rethinking}, with the ground-truth explanations, we can frame the evaluation problem as the task of top edge ranking with \textbf{Recall@N} and \textbf{Precision@N}. To be more specific, for an explanatory subgraph, the edges within the motif are positive, while the others are negative. To this end, recall and precision can be adopted as the evaluation protocols. More formally, Recall@N $= \mathbb{E}_{\mathcal{G}} ( \vert \mathcal{G}_{S} \cap \mathcal{G}^{*}_{S} \vert / \vert \mathcal{G}^{*}_{S} \vert ) $ and   Precision@N $= \mathbb{E}_{\mathcal{G}} ( \vert \mathcal{G}_{S} \cap \mathcal{G}^{*}_{S} \vert  / \vert \mathcal{G}_{S} \vert   ) $ where $\mathcal{G}_{S}$ is composed of the top-N edges and $\mathcal{G}^{*}_{S}$ is the ground-truth explanations.


\subsection{Quantitative Experiments}

\paragraph{GNN Explanation Performance.} To investigate the explainability of our method, we conducted experiments on four datasets. As reported in Table~\ref{xai-table},  our method achieves state-of-the-art performance on the explanation benchmarks, outperforming existing methods in most cases. Specifically, our method achieves significant improvements over the strongest baseline \textit{w.r.t.} Recall@5 by 0.328\% and 1.261\%, Precision@5 by 4.623\% and 16.53\% on the BA3-Motif and MUTAG datasets, respectively. 
These improvements should be attributed to the advantage of the subgraph retrieval module in identifying the clear subgraph patterns with the global view of the dataset, where
the large improvement of Precision@5  indicates our strong ability to distinguish non-critical substructures from the compression of critical subgraphs by our causal module.

In complex real-world scenarios (Benzene and MutagenicityV2), our method consistently shows state-of-the-art performances.
On the Benzene dataset, our method achieves the improvements of 5.373\% in Recall@5 and 2.361\% in Precision@5. On the more complicated MutagenicityV2 dataset, our method expanded the improvement margins with 8.064\% in Recall@5 and 35.24\% in Precision@5.
This is in line with our expectations since RC-GNN could capture the explanatory patterns shared by graph instances in the same category, As a result, our method not only focuses on retrieving rich and meaningful subgraphs, but also compresses them through the causal module.


\paragraph{GNN Prediction Performance.} To indicate the ability of our method to increment GNN predictions, we performed experiments on two explanatory datasets, and supplemented the HIV and BBBP datasets in the Open Graph benchmark~\cite{hu2020open}. We don't show BA3-Motif and Benzene results, as all methods achieved testing accuracy close to 100\%. As shown in Table~\ref{gnn-table}, our method RC-GNN outperforms other augmentation baselines with explainers consistently by a large margin.  DropNode,  a data augmentation technique, yields a marginal improvement in the GNN performance. PGDrop and MatchDrop, which sample the subgraph through corresponding explainers (PGExplainer and MatchExplainer), effectively improve GNN performance on the datasets with clear subgraph patterns (MUTAG). However, on the molecular benchmarks with diverse subgraph patterns, PGDrop and MatchDrop have a negative impact on the GNN performance. This is likely because these two explainers don't participate in GNN training, and is difficult to explore the diverse and complex explanatory subgraphs. 

When compared with existing interpretable GNN models (GIB, CAL, and DisC), RC-GNN outperforms the strongest baselines (CAL) by an average of 5.46\%. Notably, the interpretable GNNs, although improving over MatchDrop,  often decrease their interpretability for improving the predictions of the target label. We have further conducted the ablation studies, and the removal of the subgraph retrieval or the causal module led to a sharp drop in performance  (Appendix~\ref{sec-sup-abl}). These results indicate the effectiveness of incorporating the subgraph retrieval into the causal theory on GNN.

\begin{table}[t]
\caption{Explainability performance for our method and other baselines on Molecular Property Cliffs.}
\label{cliffs-table}
\begin{center}
\resizebox{0.95\linewidth}{!}
{
\begin{tabular}{lcccc}
\toprule

\multirow{3}{*}{Benchmark} & \multicolumn{4}{c}{CYP3A4} \\

& ACC-AUC & Rec@5 & Prec@5 & ACC \\& & & & (Graph-level)\\

\midrule
GradCAM  & 0.726 & 0.018 & 0.014 & - \\
SAExplainer  & 0.616 & 0.074 & 0.064 & - \\
GNNExplainer  & 0.851 & 0.128 & 0.149 & - \\
PGExplainer  & 0.591 & 0.107 & 0.093 & 0.780 \\
PGMExplainer  & 0.633 & 0.167 & 0.121 & - \\
ReFine  & 0.785 & 0.235 & 0.207  & - \\
MatchExplainer  & 0.669 & 0.134 & 0.113 & 0.798 \\

\midrule
RC-Explainer  & \textbf{0.796} & \textbf{0.344} & \textbf{0.303}  & \textbf{0.830} \\
Relative Impro. & 1.401 & 31.14 & 46.38 & 4.010  \\

\bottomrule
\end{tabular}
}
\end{center}
\vskip -0.1in
\end{table}

\paragraph{Exploration on the Molecular Property Cliff Benchmark.} 
Our method was further evaluated on the molecular property cliff benchmark, a more challenging dataset. The dataset contains key subgraphs that are more diverse and complicated, which requires correctly finding the key substructures for molecular activity. 
As shown in Table~\ref{cliffs-table}, all methods have decreased significantly in terms of precision and recall, while our method RC-GNN significantly outperforms all baselines across all metrics. Specifically, RC-GNN improves over the strongest baseline with the Recall@5 by 31.14\%, Precision@5 by 46.38\%, and classification Accuracy by 4.01\%. The large improvements by RC-GNN are mostly contributed from our retrieval-based causal learning, where the explanations not only own global understanding of model workings (i.e. the possible key substructures of activity), but also account for local insights on specific graphs (i.e. the causal effect of subgraph). These illustrate the importance of our semi-parametric training paradigm to incorporate subgraph retrieval with causal learning.

\subsection{Visualization Analysis}

\paragraph{Visualization of explanations.} To better illustrate the extracted causal and trivial subgraphs by our method, we visualize the explanations of MatchExplainer and our method on MUTAG and CYP3A4. 

As shown in Fig. 2 and Appendix~\ref{sup-vis}, on the MUTAG dataset, MatchExplainer partly identifies the actual substructure but falsely focuses on other substructures such as hydrogen atoms. The correct identification by our method should be attributed to the minimization of the mutual information between the input graph and the compressed subgraph.  On the property cliff dataset, MatchExplainer, depending on the subgraph matching paradigm, is misguided by the frequent subgraph on local information. Our method,  because of the retrieval paradigm, could summarize the patterns across the whole dataset and has a global view of the target model.



\paragraph{Projection of disentangled representation. } Fig. 3 shows the projection of our learned latent vectors $H_{\mathcal{G}}$, $H_{\mathcal{G}_c}$, and $H_{\mathcal{G}_t}$ respectively extracted from the pretrained GNN $f$, causal GNN $f_{c}$ and trivial GNN $f_{t}$   on the MUTAG and MutagenicityV2 datasets. Obviously, $H_{\mathcal{G}_c}$ are well clustered according to the graph labels,  while the trivial features $H_{\mathcal{G}_t}$ are mixed with graph labels.  $H_{\mathcal{G}} = H_{\mathcal{G}_c \cup \mathcal{G}_t} $, containing both causal features and trivial information, are clustered worse than $H_{\mathcal{G}_c}$ but better than $H_{\mathcal{G}_t}$. These results indicate that our method successfully incorporates the causal representations for the improvements of GNN.

\begin{figure}[t]
\label{fig-case}
\begin{center}
\centerline{\includegraphics[width=\columnwidth]{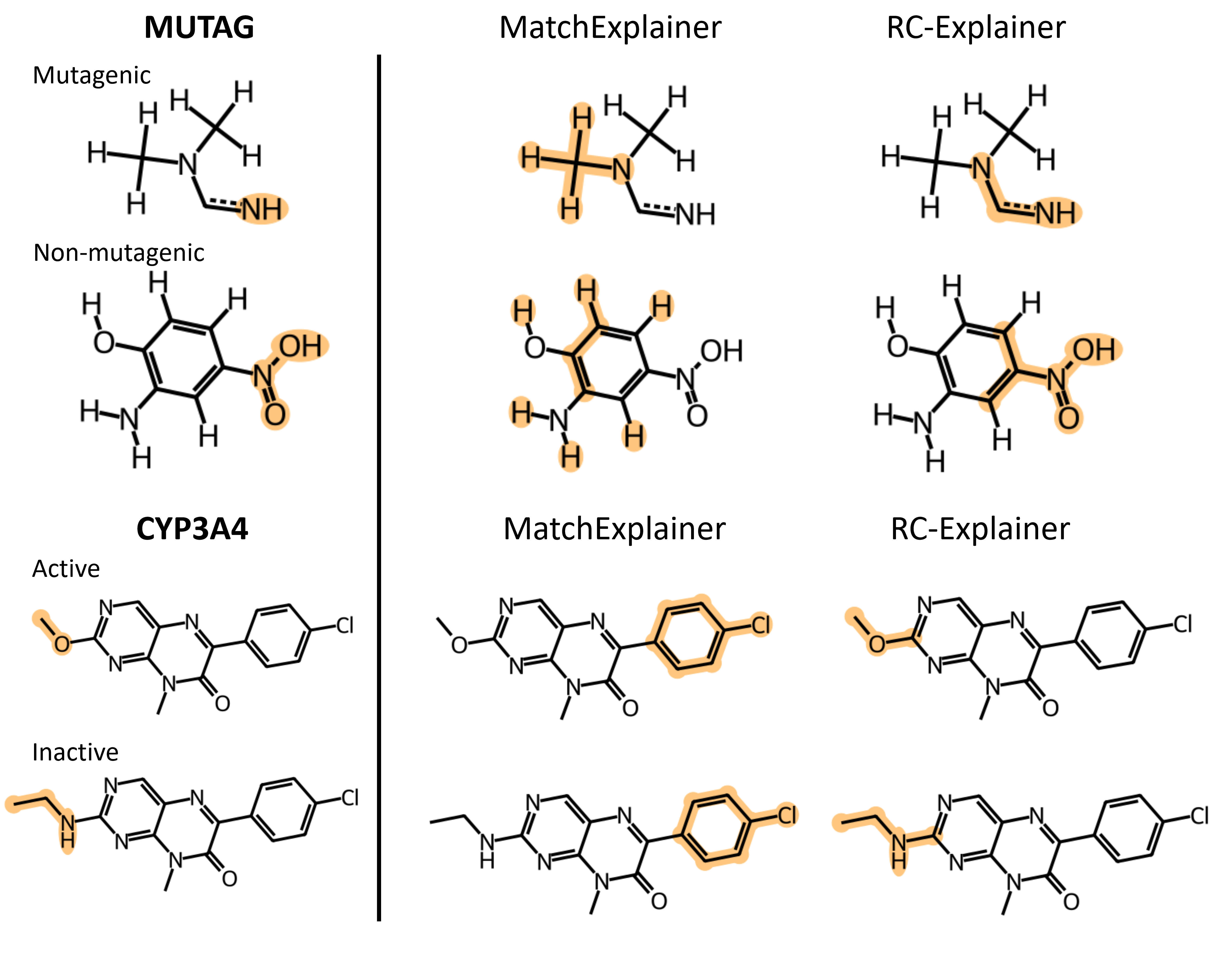}}
\caption{Explanatory subgraphs for MUTAG and CYP3A4.}
\end{center}
\vskip -0.2in
\end{figure}

\begin{figure}[t]
\label{fig-tsne}
\begin{center}
\centerline{\includegraphics[width=\columnwidth]{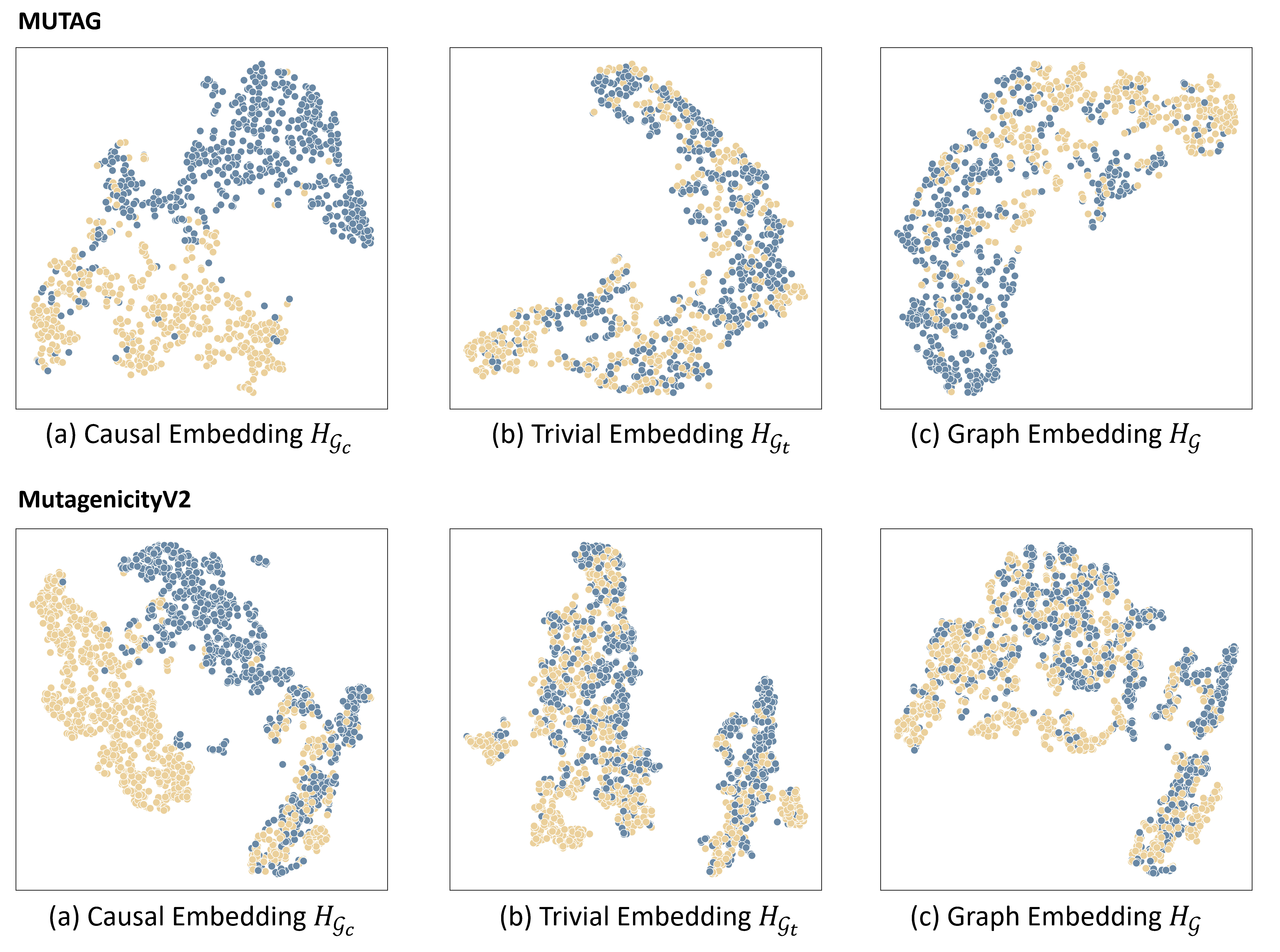}}
\caption{Visualization of $H_{\mathcal{G}_c}$, $H_{\mathcal{G}_t}$ and 
$H_{\mathcal{G}}$ on MUTAG and MutagenicityV2 by t-SNE~\cite{van2008visualizing} with colors labeled by the graph labels.}
\end{center}
\vskip -0.2in
\end{figure}

\section{Conclusion}
This paper proposes a promising interpretable GNN framework called RC-GNN. Distinct from the popular trend of post-hoc explainers and transparent GNNs, the framework incorporates retrieval-based causal learning with the GIB theory, providing both reliable explanations and generalizable predictions. Experiments convincingly demonstrate the efficacy of the interpretability on explanation benchmarks, and in turn contribute to enhancing the performance of graph classification. This study is expected to push the frontier of retrieval-based methods to explain deep learning models.

\section*{Impact Statements}
This paper presents work whose goal is to advance the field of explainable artificial intelligence (XAI). There are many potential societal consequences of our work, none of which we feel must be specifically highlighted here.






\nocite{langley00}

\bibliography{example_paper}

\begin{thebibliography}{44}
\providecommand{\natexlab}[1]{#1}
\providecommand{\url}[1]{\texttt{#1}}
\expandafter\ifx\csname urlstyle\endcsname\relax
  \providecommand{\doi}[1]{doi: #1}\else
  \providecommand{\doi}{doi: \begingroup \urlstyle{rm}\Url}\fi

\bibitem[Achille \& Soatto(2018)Achille and Soatto]{achille2018emergence}
Achille, A. and Soatto, S.
\newblock Emergence of invariance and disentanglement in deep representations.
\newblock \emph{The Journal of Machine Learning Research}, 19\penalty0 (1):\penalty0 1947--1980, 2018.

\bibitem[Agarwal et~al.(2023)Agarwal, Queen, Lakkaraju, and Zitnik]{agarwal2023evaluating}
Agarwal, C., Queen, O., Lakkaraju, H., and Zitnik, M.
\newblock Evaluating explainability for graph neural networks.
\newblock \emph{Scientific Data}, 10\penalty0 (1):\penalty0 144, 2023.

\bibitem[Amann et~al.(2020)Amann, Blasimme, Vayena, Frey, Madai, and Consortium]{amann2020explainability}
Amann, J., Blasimme, A., Vayena, E., Frey, D., Madai, V.~I., and Consortium, P.
\newblock Explainability for artificial intelligence in healthcare: a multidisciplinary perspective.
\newblock \emph{BMC medical informatics and decision making}, 20:\penalty0 1--9, 2020.

\bibitem[Baldassarre \& Azizpour(2019)Baldassarre and Azizpour]{baldassarre2019explainability}
Baldassarre, F. and Azizpour, H.
\newblock Explainability techniques for graph convolutional networks.
\newblock \emph{arXiv preprint arXiv:1905.13686}, 2019.

\bibitem[Chen et~al.(2022)Chen, Zhang, Bian, Yang, Kaili, Xie, Liu, Han, and Cheng]{chen2022learning}
Chen, Y., Zhang, Y., Bian, Y., Yang, H., Kaili, M., Xie, B., Liu, T., Han, B., and Cheng, J.
\newblock Learning causally invariant representations for out-of-distribution generalization on graphs.
\newblock \emph{Advances in Neural Information Processing Systems}, 35:\penalty0 22131--22148, 2022.

\bibitem[Fan et~al.(2022)Fan, Wang, Mo, Shi, and Tang]{fan2022debiasing}
Fan, S., Wang, X., Mo, Y., Shi, C., and Tang, J.
\newblock Debiasing graph neural networks via learning disentangled causal substructure.
\newblock \emph{Advances in Neural Information Processing Systems}, 35:\penalty0 24934--24946, 2022.

\bibitem[Fan et~al.(2019)Fan, Ma, Li, He, Zhao, Tang, and Yin]{fan2019graph}
Fan, W., Ma, Y., Li, Q., He, Y., Zhao, E., Tang, J., and Yin, D.
\newblock Graph neural networks for social recommendation.
\newblock In \emph{The world wide web conference}, pp.\  417--426, 2019.

\bibitem[Gilmer et~al.(2017)Gilmer, Schoenholz, Riley, Vinyals, and Dahl]{gilmer2017neural}
Gilmer, J., Schoenholz, S.~S., Riley, P.~F., Vinyals, O., and Dahl, G.~E.
\newblock Neural message passing for quantum chemistry.
\newblock In \emph{International conference on machine learning}, pp.\  1263--1272. PMLR, 2017.

\bibitem[Hamilton et~al.(2017)Hamilton, Ying, and Leskovec]{hamilton2017inductive}
Hamilton, W., Ying, Z., and Leskovec, J.
\newblock Inductive representation learning on large graphs.
\newblock \emph{Advances in neural information processing systems}, 30, 2017.

\bibitem[Hu et~al.(2020)Hu, Fey, Zitnik, Dong, Ren, Liu, Catasta, and Leskovec]{hu2020open}
Hu, W., Fey, M., Zitnik, M., Dong, Y., Ren, H., Liu, B., Catasta, M., and Leskovec, J.
\newblock Open graph benchmark: Datasets for machine learning on graphs.
\newblock \emph{Advances in neural information processing systems}, 33:\penalty0 22118--22133, 2020.

\bibitem[Huang et~al.(2022)Huang, Yamada, Tian, Singh, and Chang]{huang2022graphlime}
Huang, Q., Yamada, M., Tian, Y., Singh, D., and Chang, Y.
\newblock Graphlime: Local interpretable model explanations for graph neural networks.
\newblock \emph{IEEE Transactions on Knowledge and Data Engineering}, 2022.

\bibitem[Hussain \& Rea(2010)Hussain and Rea]{hussain2010computationally}
Hussain, J. and Rea, C.
\newblock Computationally efficient algorithm to identify matched molecular pairs (mmps) in large data sets.
\newblock \emph{Journal of chemical information and modeling}, 50\penalty0 (3):\penalty0 339--348, 2010.

\bibitem[Kipf \& Welling(2017)Kipf and Welling]{kipf2017semisupervised}
Kipf, T.~N. and Welling, M.
\newblock Semi-supervised classification with graph convolutional networks.
\newblock In \emph{International Conference on Learning Representations}, 2017.
\newblock URL \url{https://openreview.net/forum?id=SJU4ayYgl}.

\bibitem[Langley(2000)]{langley00}
Langley, P.
\newblock Crafting papers on machine learning.
\newblock In Langley, P. (ed.), \emph{Proceedings of the 17th International Conference on Machine Learning (ICML 2000)}, pp.\  1207--1216, Stanford, CA, 2000. Morgan Kaufmann.

\bibitem[Lin et~al.(2021)Lin, Lan, and Li]{lin2021generative}
Lin, W., Lan, H., and Li, B.
\newblock Generative causal explanations for graph neural networks.
\newblock In \emph{International Conference on Machine Learning}, pp.\  6666--6679. PMLR, 2021.

\bibitem[Lucic et~al.(2022)Lucic, Ter~Hoeve, Tolomei, De~Rijke, and Silvestri]{lucic2022cf}
Lucic, A., Ter~Hoeve, M.~A., Tolomei, G., De~Rijke, M., and Silvestri, F.
\newblock Cf-gnnexplainer: Counterfactual explanations for graph neural networks.
\newblock In \emph{International Conference on Artificial Intelligence and Statistics}, pp.\  4499--4511. PMLR, 2022.

\bibitem[Luo et~al.(2020)Luo, Cheng, Xu, Yu, Zong, Chen, and Zhang]{luo2020parameterized}
Luo, D., Cheng, W., Xu, D., Yu, W., Zong, B., Chen, H., and Zhang, X.
\newblock Parameterized explainer for graph neural network.
\newblock \emph{Advances in neural information processing systems}, 33:\penalty0 19620--19631, 2020.

\bibitem[Pei et~al.(2020)Pei, Yu, and Tian]{pei2020amalnet}
Pei, X., Yu, L., and Tian, S.
\newblock Amalnet: A deep learning framework based on graph convolutional networks for malware detection.
\newblock \emph{Computers \& Security}, 93:\penalty0 101792, 2020.

\bibitem[Pope et~al.(2019)Pope, Kolouri, Rostami, Martin, and Hoffmann]{pope2019explainability}
Pope, P.~E., Kolouri, S., Rostami, M., Martin, C.~E., and Hoffmann, H.
\newblock Explainability methods for graph convolutional neural networks.
\newblock In \emph{Proceedings of the IEEE/CVF conference on computer vision and pattern recognition}, pp.\  10772--10781, 2019.

\bibitem[Ranjan et~al.(2020)Ranjan, Sanyal, and Talukdar]{ranjan2020asap}
Ranjan, E., Sanyal, S., and Talukdar, P.
\newblock Asap: Adaptive structure aware pooling for learning hierarchical graph representations.
\newblock In \emph{Proceedings of the AAAI Conference on Artificial Intelligence}, volume~34, pp.\  5470--5477, 2020.

\bibitem[Rao et~al.(2022)Rao, Zheng, Lu, and Yang]{rao2022quantitative}
Rao, J., Zheng, S., Lu, Y., and Yang, Y.
\newblock Quantitative evaluation of explainable graph neural networks for molecular property prediction.
\newblock \emph{Patterns}, 3\penalty0 (12), 2022.

\bibitem[Sanchez-Lengeling et~al.(2020)Sanchez-Lengeling, Wei, Lee, Reif, Wang, Qian, McCloskey, Colwell, and Wiltschko]{sanchez2020evaluating}
Sanchez-Lengeling, B., Wei, J., Lee, B., Reif, E., Wang, P., Qian, W., McCloskey, K., Colwell, L., and Wiltschko, A.
\newblock Evaluating attribution for graph neural networks.
\newblock \emph{Advances in neural information processing systems}, 33:\penalty0 5898--5910, 2020.

\bibitem[Schlichtkrull et~al.(2018)Schlichtkrull, Kipf, Bloem, Van Den~Berg, Titov, and Welling]{schlichtkrull2018modeling}
Schlichtkrull, M., Kipf, T.~N., Bloem, P., Van Den~Berg, R., Titov, I., and Welling, M.
\newblock Modeling relational data with graph convolutional networks.
\newblock In \emph{The Semantic Web: 15th International Conference, ESWC 2018, Heraklion, Crete, Greece, June 3--7, 2018, Proceedings 15}, pp.\  593--607. Springer, 2018.

\bibitem[Seo et~al.(2023)Seo, Kim, and Park]{seo2023interpretable}
Seo, S., Kim, S., and Park, C.
\newblock Interpretable prototype-based graph information bottleneck.
\newblock In \emph{Thirty-seventh Conference on Neural Information Processing Systems}, 2023.
\newblock URL \url{https://openreview.net/forum?id=icWwBKyVMs}.

\bibitem[Song et~al.(2020)Song, Zheng, Niu, Fu, Lu, and Yang]{song2020communicative}
Song, Y., Zheng, S., Niu, Z., Fu, Z.-H., Lu, Y., and Yang, Y.
\newblock Communicative representation learning on attributed molecular graphs.
\newblock In \emph{IJCAI}, volume 2020, pp.\  2831--2838, 2020.

\bibitem[Sterling \& Irwin(2015)Sterling and Irwin]{sterling2015zinc}
Sterling, T. and Irwin, J.~J.
\newblock Zinc 15--ligand discovery for everyone.
\newblock \emph{Journal of chemical information and modeling}, 55\penalty0 (11):\penalty0 2324--2337, 2015.

\bibitem[Sui et~al.(2022)Sui, Wang, Wu, Lin, He, and Chua]{sui2022causal}
Sui, Y., Wang, X., Wu, J., Lin, M., He, X., and Chua, T.-S.
\newblock Causal attention for interpretable and generalizable graph classification.
\newblock In \emph{Proceedings of the 28th ACM SIGKDD Conference on Knowledge Discovery and Data Mining}, pp.\  1696--1705, 2022.

\bibitem[Sushko et~al.(2012)Sushko, Salmina, Potemkin, Poda, and Tetko]{sushko2012toxalerts}
Sushko, I., Salmina, E., Potemkin, V.~A., Poda, G., and Tetko, I.~V.
\newblock Toxalerts: a web server of structural alerts for toxic chemicals and compounds with potential adverse reactions, 2012.

\bibitem[Van~der Maaten \& Hinton(2008)Van~der Maaten and Hinton]{van2008visualizing}
Van~der Maaten, L. and Hinton, G.
\newblock Visualizing data using t-sne.
\newblock \emph{Journal of machine learning research}, 9\penalty0 (11), 2008.

\bibitem[Veith et~al.(2009)Veith, Southall, Huang, James, Fayne, Artemenko, Shen, Inglese, Austin, Lloyd, et~al.]{veith2009comprehensive}
Veith, H., Southall, N., Huang, R., James, T., Fayne, D., Artemenko, N., Shen, M., Inglese, J., Austin, C.~P., Lloyd, D.~G., et~al.
\newblock Comprehensive characterization of cytochrome p450 isozyme selectivity across chemical libraries.
\newblock \emph{Nature biotechnology}, 27\penalty0 (11):\penalty0 1050--1055, 2009.

\bibitem[Veličković et~al.(2018)Veličković, Cucurull, Casanova, Romero, Liò, and Bengio]{gat2018graph}
Veličković, P., Cucurull, G., Casanova, A., Romero, A., Liò, P., and Bengio, Y.
\newblock Graph attention networks.
\newblock In \emph{International Conference on Learning Representations}, 2018.
\newblock URL \url{https://openreview.net/forum?id=rJXMpikCZ}.

\bibitem[Vu \& Thai(2020)Vu and Thai]{vu2020pgm}
Vu, M. and Thai, M.~T.
\newblock Pgm-explainer: Probabilistic graphical model explanations for graph neural networks.
\newblock \emph{Advances in neural information processing systems}, 33:\penalty0 12225--12235, 2020.

\bibitem[Wang et~al.(2021)Wang, Wu, Zhang, He, and Chua]{wang2021towards}
Wang, X., Wu, Y., Zhang, A., He, X., and Chua, T.-S.
\newblock Towards multi-grained explainability for graph neural networks.
\newblock \emph{Advances in Neural Information Processing Systems}, 34:\penalty0 18446--18458, 2021.

\bibitem[Wu et~al.(2023)Wu, Li, Jin, Jiang, Radev, Niu, and Li]{wu2023rethinking}
Wu, F., Li, S., Jin, X., Jiang, Y., Radev, D., Niu, Z., and Li, S.~Z.
\newblock Rethinking explaining graph neural networks via non-parametric subgraph matching.
\newblock In \emph{International Conference on Machine Learning}, pp.\  37511--37523. PMLR, 2023.

\bibitem[Wu et~al.(2020)Wu, Ren, Li, and Leskovec]{wu2020graph}
Wu, T., Ren, H., Li, P., and Leskovec, J.
\newblock Graph information bottleneck.
\newblock \emph{Advances in Neural Information Processing Systems}, 33:\penalty0 20437--20448, 2020.

\bibitem[Xie \& Grossman(2018)Xie and Grossman]{xie2018crystal}
Xie, T. and Grossman, J.~C.
\newblock Crystal graph convolutional neural networks for an accurate and interpretable prediction of material properties.
\newblock \emph{Physical review letters}, 120\penalty0 (14):\penalty0 145301, 2018.

\bibitem[Xu et~al.(2019)Xu, Hu, Leskovec, and Jegelka]{xu2018how}
Xu, K., Hu, W., Leskovec, J., and Jegelka, S.
\newblock How powerful are graph neural networks?
\newblock In \emph{International Conference on Learning Representations}, 2019.
\newblock URL \url{https://openreview.net/forum?id=ryGs6iA5Km}.

\bibitem[Ying et~al.(2019)Ying, Bourgeois, You, Zitnik, and Leskovec]{ying2019gnnexplainer}
Ying, Z., Bourgeois, D., You, J., Zitnik, M., and Leskovec, J.
\newblock Gnnexplainer: Generating explanations for graph neural networks.
\newblock \emph{Advances in neural information processing systems}, 32, 2019.

\bibitem[You et~al.(2020)You, Chen, Sui, Chen, Wang, and Shen]{you2020graph}
You, Y., Chen, T., Sui, Y., Chen, T., Wang, Z., and Shen, Y.
\newblock Graph contrastive learning with augmentations.
\newblock \emph{Advances in neural information processing systems}, 33:\penalty0 5812--5823, 2020.

\bibitem[Yu et~al.(2021)Yu, Xu, Rong, Bian, Huang, and He]{yu2021graph}
Yu, J., Xu, T., Rong, Y., Bian, Y., Huang, J., and He, R.
\newblock Graph information bottleneck for subgraph recognition.
\newblock In \emph{International Conference on Learning Representations}, 2021.
\newblock URL \url{https://openreview.net/forum?id=bM4Iqfg8M2k}.

\bibitem[Yuan et~al.(2020)Yuan, Tang, Hu, and Ji]{yuan2020xgnn}
Yuan, H., Tang, J., Hu, X., and Ji, S.
\newblock Xgnn: Towards model-level explanations of graph neural networks.
\newblock In \emph{Proceedings of the 26th ACM SIGKDD International Conference on Knowledge Discovery \& Data Mining}, pp.\  430--438, 2020.

\bibitem[Yuan et~al.(2021)Yuan, Yu, Wang, Li, and Ji]{yuan2021explainability}
Yuan, H., Yu, H., Wang, J., Li, K., and Ji, S.
\newblock On explainability of graph neural networks via subgraph explorations.
\newblock In \emph{International conference on machine learning}, pp.\  12241--12252. PMLR, 2021.

\bibitem[Zhang et~al.(2018)Zhang, Cui, Neumann, and Chen]{zhang2018end}
Zhang, M., Cui, Z., Neumann, M., and Chen, Y.
\newblock An end-to-end deep learning architecture for graph classification.
\newblock In \emph{Proceedings of the AAAI conference on artificial intelligence}, volume~32, 2018.

\bibitem[Zheng et~al.(2021)Zheng, Rao, Song, Zhang, Xiao, Fang, Yang, and Niu]{zheng2021pharmkg}
Zheng, S., Rao, J., Song, Y., Zhang, J., Xiao, X., Fang, E.~F., Yang, Y., and Niu, Z.
\newblock Pharmkg: a dedicated knowledge graph benchmark for bomedical data mining.
\newblock \emph{Briefings in bioinformatics}, 22\penalty0 (4):\penalty0 bbaa344, 2021.

\end{thebibliography}
\bibliographystyle{icml2024}













\newpage

\appendix
\onecolumn



\section{Preliminaries and Methodology}
\label{sup-meth}



\subsection{Further Theoretical Analysis}

According to the theory of Graph Information Bottlenecks (GIB), the explanation problem can be formulated as the following optimization problem based on the mutual information:

\begin{equation}
\label{eq:sup-gib}
    \mathop{\rm max} \limits_{\mathcal{G}_{S} \subset \mathcal{G} } \left[  I (Y; \mathcal{G}_{S}) - \beta I(\mathcal{G}; \mathcal{G}_{S}) \right]
\end{equation}

The first term maximizes the mutual information between the graph label and the compressed subgraph,  ensuring that the compressed subgraph contains maximal information for predicting the graph label. The second term minimizes the mutual information between the input graph and the compressed subgraph,  ensuring that the compressed subgraph contains minimal information about the input graph.

For the first term, we can reformulate it as:

\begin{equation}
\label{eq:gib-re}
\begin{aligned}
     \mathop{\rm max} \limits_{\mathcal{G}_{S} \subset \mathcal{G} } I (Y; \mathcal{G}_{S}) 
     = \mathop{\rm max} \limits_{\mathcal{G}_{S} \subset \mathcal{G} } 
     \mathbb{E}_{(Y, \mathcal{G}_{S})} \left[ 
     {\rm log} {\frac{p (Y, \mathcal{G}_{S})}{ p(Y) p(\mathcal{G}_{S}) }} 
     \right] \left( s.t. \vert \mathcal{G}_{S} \vert < K \right)
\end{aligned}
\end{equation}
where K is the subgraph size on $\mathcal{G}$ for a compact explanation.



We allow the involvement of the graph $\mathcal{G}^{'}$ in Eq.~\ref{eq:gib-re}, which is from the set of graphs in the same class that typically share partial common motif patterns. Therefore, the optimization problem can be approximated as:

\begin{equation}
\begin{aligned}
\label{eq:involvement}
     \mathbb{E}_{(Y, \mathcal{G}_{S})} \left[ 
     {\rm log} {\frac{p (Y, \mathcal{G}_{S})}{ p(Y) p(\mathcal{G}_{S}) }}
     \right] &= 
     \mathbb{E}_{(Y, \mathcal{G}_{S}, \mathcal{G}^{'})} \left[ 
     {\rm log} {\frac{p (Y, \mathcal{G}_{S}, \mathcal{G}^{'})}{ p(Y) p(\mathcal{G}_{S}, \mathcal{G}^{'}) }}
     \right]  + \mathbb{E}_{(Y, \mathcal{G}_{S}, \mathcal{G}^{'})} \left[ 
     {\rm log} {\frac{p (Y, \mathcal{G}_{S}) p(\mathcal{G}_{S}, \mathcal{G}^{'})}{ p(Y, \mathcal{G}_{S}, \mathcal{G}^{'}) p(\mathcal{G}_{S}) }}
     \right] \\
     &= \mathbb{E}_{(Y, \mathcal{G}_{S}, \mathcal{G}^{'})} \left[ 
     {\rm log} {\frac{p (Y, \mathcal{G}_{S}, \mathcal{G}^{'})}{ p(Y) p(\mathcal{G}_{S}, \mathcal{G}^{'}) }}
     \right] -  \mathbb{E}_{(Y, \mathcal{G}_{S}, \mathcal{G}^{'})} \left[ 
     {\rm log} {\frac{p (Y, \mathcal{G}_{S}, \mathcal{G}^{'}) p(\mathcal{G}_{S} )}{ p(Y, \mathcal{G}_{S}) p(\mathcal{G}_{S}, \mathcal{G}^{'}) }}
     \right] \\
     &= \mathbb{E}_{(Y, \mathcal{G}_{S}, \mathcal{G}^{'})} \left[ 
     {\rm log} {\frac{p (Y, \mathcal{G}_{S}, \mathcal{G}^{'})}{ p(Y) p(\mathcal{G}_{S}, \mathcal{G}^{'}) }}
     \right] -  \mathbb{E}_{(Y, \mathcal{G}_{S}, \mathcal{G}^{'})} \left[ 
     {\rm log} {\frac{p (Y, \mathcal{G}^{'} \vert \mathcal{G}_{S})}{ p(Y \vert \mathcal{G}_{S}) p(\mathcal{G}^{'} \vert \mathcal{G}_{S}) }}
     \right] \\
     &= I(Y; \mathcal{G}_{S}, \mathcal{G}^{'}) - I(Y; \mathcal{G}^{'} \vert \mathcal{G}_{S}) 
\end{aligned}
\end{equation}

Following \citet{seo2023interpretable}, we decompose the second term of Eq.~\ref{eq:involvement} into the sum of two terms based on the chain rule of mutual information as follows:

\begin{equation}
\label{eq-decom}
    I(Y; \mathcal{G}_{S}, \mathcal{G}^{'}) - I(Y; \mathcal{G}^{'} \vert \mathcal{G}_{S}) = I(Y; \mathcal{G}_{S}, \mathcal{G}^{'} ) - I(\mathcal{G}^{'}; Y, \mathcal{G}_{S}) + I(\mathcal{G}^{'}, \mathcal{G}_{S})  
\end{equation}

The first term $I(Y; \mathcal{G}_{S}, \mathcal{G}^{'})$ in Eq.~\ref{eq-decom} is to maximize the mutual information between $Y$ and the joint variables $(\mathcal{G}_{S}, \mathcal{G}^{'})$. The second term is to minimize the mutual information between $\mathcal{G}^{'}$ and the joint variables $(Y, \mathcal{G}_{S})$. The third term is to guarantee the mutual information between $\mathcal{G}^{'}$ and $\mathcal{G}_{S}$. Since the second term eliminates the information about $Y$ related to $\mathcal{G}_{S}$ from $\mathcal{G}^{'}$ and we aimed to ensure the interpretability of the $\mathcal{G}_{S}$ in $\mathcal{G}$, we excluded the second term during training, and only considered the objective of $I(Y; \mathcal{G}_{S}, \mathcal{G}^{'} ) + I(\mathcal{G}^{'}, \mathcal{G}_{S})$.

Therefore, our target is equivalent to optimizing the maximization of the mutual information between the subgraph $\mathcal{G}_{S}$ and the explanatory subgraph $\mathcal{G}^{'}_{S}$ in graph $\mathcal{G}^{'}$ from the candidate set $\mathcal{D}$. The candidate set is composed of the graph $\mathcal{G}^{'}$ that shares the same category with $\mathcal{G}_{S}$ to guarantee the mutual information between $Y$ and $(\mathcal{G}_{S}, \mathcal{G}^{'})$ during our training process. By optimizing the cross-entropy loss for a categorical $Y$ with $\mathcal{G}_{S}$, it can be formulated as:

\begin{equation}
    \mathop{\rm max}\limits_{\mathcal{G}^{'} \subset \mathcal{D} } \left[ 
        \mathop{\rm max} \limits_{\mathcal{G}_{S} \subset \mathcal{G}, \mathcal{G}^{'}_{S} \subset \mathcal{G}^{'} } I (\mathcal{G}^{'}_{S}, \mathcal{G}_{S}) 
    \right]
\end{equation}

Then we consider the minimization of $I(\mathcal{G}, \mathcal{G}_{S}) $  to ensure that the compressed subgraph contains minimal information about the input graph, which is the second term of Eq.~\ref{eq:sup-gib}. 
We utilized the causal theory on GNN to consider the $\mathcal{G}_{S}$ in graph $\mathcal{G}$ as the causal graph and the $\mathcal{G}_{t} = \mathcal{G} - \mathcal{G}_{S}$ as the trivial graph. We could achieve the graph representation by eliminating the backdoor path with the causal theory. We can exploit the do-calculus on the variable to remove the backdoor path by estimating:
\begin{equation}
\begin{aligned}
\label{eq:sup-backdoor}
    P(\mathcal{G} \vert  do(\mathcal{G}_{S}) ) &= P({\mathcal{G} \vert  \mathcal{G}_{S}}) \\
        & = \sum\limits_{\mathcal{G}_t \subset \mathcal{D}_{t}} { P(\mathcal{G} \vert  \mathcal{G}_{S}, \mathcal{G}_t) P(\mathcal{G}_t \vert \mathcal{G}_{S})}  \\ 
        &= \sum\limits_{\mathcal{G}_t \subset \mathcal{D}_{t}} { P(\mathcal{G} \vert  \mathcal{G}_{S}, \mathcal{G}_t) P(\mathcal{G}_t)} \\  %
        &= \sum\limits_{\mathcal{G}_t \subset \mathcal{D}_{t}} { P(\mathcal{G} \vert  \mathcal{G}_{S}, \mathcal{G}_t) P(\mathcal{G}_t)}
\end{aligned}
\end{equation}
where $\mathcal{D}_{t}$ denotes the confounder set, $P(\mathcal{G} \vert  \mathcal{G}_{c}, \mathcal{G}_t)$ represents the conditional probability given the causal subgraph $\mathcal{G}_{S}$ and confounder $\mathcal{G}_t$, and $P(\mathcal{G}_t)$ is the prior probability of the confounder. Eq. (\ref{eq:backdoor}) is called backdoor adjustment.


Recent studies on contrastive learning have proven that minimizing contrastive loss is equivalent to maximizing the mutual information between two variables~\cite{you2020graph}.  
Hence, we minimize the contrastive loss $\mathcal{L}$ between the causal subgraph and trivial subgraph for implementing Eq.~\ref{eq:sup-backdoor}. 
More precisely, we consider $\mathcal{G}_{S}$ and $\mathcal{G}_{t}$ in the same graph as a positive pair, and the typical contrastive loss is defined as follows:

\begin{equation}
    \mathcal{L} = \frac{1}{n} \sum\limits_{i=1}^{n}
    {\rm log} \frac{
        \sum\limits_{j: H_{\mathcal{G}_{t}^{j}} \in \mathbb{P} } {\rm exp} \left( { ( H_{\mathcal{G}_{S}^{}}, H_{\mathcal{G}_{t}^{j}} ) / \tau } \right)
    }{
        \sum\limits_{k: H_{\mathcal{G}_{t}^{k}} \not\in \mathbb{P}  } {\rm exp} \left( { ( H_{\mathcal{G}_{S}^{k}}, H_{\mathcal{G}_{t}^{k}} ) / \tau } \right)
    }
\end{equation}
where $\tau$ is the temperature hyperparameter, $n$ denotes the number of graphs in a batch, and $j$ and $k$ indicate indices of positive and negative samples, respectively. $\mathbb{P}$ is the set of positive pairs.

\subsection{Algorithm}

\begin{algorithm}[h]
   \caption{Overview of RC-GNN Training}
   \label{alg:example}
\begin{algorithmic}
   \STATE {\bfseries Input:} Training dataset $\{ (\mathcal{G}_i, y_i)\}$, the training epoch $T$
   \STATE Initialize GNN model parameters.
   \FOR{epochs $t=1$ {\bfseries to} $T$}
   \STATE Generate candidate set $\mathcal{D}$ from GNN.
   \STATE $\mathcal{G}_{S}$ $\longleftarrow$  $\mathop{\rm max}\limits_{\mathcal{G}_{S} \subset \mathcal{G} }$ $I(Y; \mathcal{G}_{S}, \mathcal{G}^{'} ) + I(\mathcal{G}^{'}, \mathcal{G}_{S})$ by retrieving from $\mathcal{D}$.
   \STATE Minimize Eq.~\ref{eq-total} via causal module.
   \STATE Update model parameters by gradient descent.
   \ENDFOR
\end{algorithmic}
\end{algorithm}

\section{Further Implementation Details}
\label{sup-impl}

\subsection{Datasets}

The statistics of four datasets are presented in Table~\ref{static-table}. Note that, we report the average number of nodes and the average number of edges over all the graphs for the real-world datasets. Table~\ref{pres-table} reports the model accuracy on four datasets, which indicates that the models to be explained are performed reasonably well.

\begin{table}[h]
\caption{Data Statistics of Five Explanation Datasets.}
\vskip 0.1in
\label{static-table}
\begin{center}
{
\begin{tabular}{lccccc}
\toprule

{Benchmark} 
& BA3-Motif & MUTAG & Benzenze & MutagenicityV2 & CYP3A4 \\

\midrule
\# Graphs  & 3000 & 4,337 & 12,000 & 6,506 & 9,126 \\
\# Nodes  & 22 & 29 & 21 & 18 & 25 \\
\# Edges  & 15 & 30 & 22 & 17 & 26 \\
\# Labels  & 3 & 2 & 2 & 2 & 2 \\
\bottomrule
\end{tabular}
}
\end{center}
\vskip -0.1in
\end{table}

\begin{table}[h]
\caption{Pretrained GNN Model Accuracy of Five Explanation Datasets.}
\vskip 0.1in
\label{pres-table}
\begin{center}
{
\begin{tabular}{lccccc}
\toprule

{Benchmark} 
& BA3-Motif & MUTAG & Benzenze & MutagenicityV2 & CYP3A4 \\

\midrule
Accuracy  & 1.000 & 0.820 & 1.000 & 0.802 & 0.783 \\
\bottomrule
\end{tabular}
}
\end{center}
\vskip -0.1in
\end{table}

\begin{table}[tb]
\caption{Data Splitting for Five Explanation Datasets.}
\vskip 0.1in
\label{split-table}
\begin{center}
{
\begin{tabular}{lcccc}
\toprule

{Benchmark} 
& \# of Training & \# of Validation & \# of Testing & \# of Explanation \\

\midrule
BA3-Motif  & 2,200 & 400 & 400 & 400 \\
MUTAG  & 3,337 & 500 & 500 & 500 \\
Benzene  & 8,400 & 1,200 & 2,400 & 2,400 \\
MutagenicityV2  & 4,404  & 800 & 1,302 & 1,302 \\
CYP3A4  & 7,310 & 812 & 903 & 101 \\
\bottomrule
\end{tabular}
}
\end{center}
\vskip -0.1in
\end{table}





\subsection{Baselines}

\begin{itemize}
[leftmargin=0in, itemindent=0.1in]
\item  \textbf{SA}~\cite{baldassarre2019explainability} directly uses the gradients of the model prediction concerning the adjacency matrix of the input graph as the importance of edges.

\item  \textbf{Grad-CAM}~\cite{pope2019explainability} uses the gradients of any target concept, such as the motif in a graph flowing into the final convolutional layer, to produce a coarse localization map highlighting the critical subgraph for predicting the concept.

\item  \textbf{GNNExplainer}~\cite{ying2019gnnexplainer} optimizes soft masks for edges and node features to maximize the mutual information between the original predictions and new predictions.

\item  \textbf{PGExplainer}~\cite{luo2020parameterized} hires a parameterized model to decide whether an edge is essential, which is trained over multiple explained instances with all edges.

\item  \textbf{PGM-Explainer}~\cite{vu2020pgm} collects the prediction change on the random node perturbations and then learns a Bayesian network from these perturbation predictions to capture the dependencies among the nodes and the prediction.

\item \textbf{ReFine}~\cite{wang2021towards} proposed a pre-training and fine-tuning framework to generate multi-grained explanations. They
have both a global understanding of model workings and local insights on specific instances. 

\item \textbf{MatchExplainer}~\cite{wu2023rethinking} proposed a non-parametric subgraph matching framework to explore explanations.



\end{itemize}

\subsection{Implementation Details}
Unless otherwise stated, all models, including GNN classification models and our explainer, are implemented using PyTorch and trained with Adam optimizer. All the experiments were performed five times on an NVIDIA GeForce RTX 4090 GPU. For the parametric explanation methods (GNNExplainer, PGExplainer, PGMExplainer, and ReFine), we have applied a grid search to tune their hyperparameters. Table~\ref{split-table} shows the detailed data splitting for model training, testing, and validation. Note that both classification models and our explanation models use the same data splitting. For those hyperparameters on our framework, we set $q$ of GCE loss as 0.7 and $\lambda_1, \lambda_2$ as 1 for all experiments. The hyperparameters in GNN models are consistent with previous studies.



\section{Additional Experimental Results}

\subsection{Ablation Study}
\label{sec-sup-abl}

We perform ablation studies to examine the effectiveness of our model. In Table~\ref{abl-table}, we show the classification performance with different variants: \textbf{RC-GNN w/o retriever} which predicted from the subgraph generator model with the causal module, \textbf{RC-GNN w/o (causal module)} which incorporates subgraph retriever and the graph augmentations without causal module, \textbf{RC-GNN w/o disentangled \& contrastive loss} which is optimized by the supervised loss between target labels and causal subgraph and \textbf{RC-GNN} is the full model. We could observe that the subgraph retriever and causal module both significantly improve the model’s performance for graph classification tasks. When substituting the subgraph retrieval with the subgraph generator model, the accuracy only outperforms MatchDrop by a small margin. When removing the causal module, the GNN performance of \textbf{RC-GNN w/o (causal module)} also drops sharply on both the MUTAG and MutagenicityV2 datasets. Causal contrastive learning helps to improve our performances further.

\begin{table}[h]
\caption{Ablation Study of RC-GNN}
\vskip 0.1in
\label{abl-table}
\begin{center}
{
\begin{tabular}{lcc}
\toprule

{Benchmark} & \multicolumn{1}{c}{MUTAG} & \multicolumn{1}{c}{MutagenicityV2} \\

\multirow{1}{*}{Metric}  & Accuracy  & Accuracy \\

\midrule
w/o (retriever) & 0.865 & 0.856 \\
w/o (causal module) & 0.882 & 0.893 \\
w/o (disentangled \& contrastive loss)  & 0.921 & 0.930 \\
RC-GNN  & \textbf{0.932} & \textbf{0.943}  \\
\bottomrule
\end{tabular}
}
\end{center}
\vskip -0.1in
\end{table}

\subsection{Visualizations}
\label{sup-vis}

\begin{figure}[hb]
\label{sup-fig-case}
\begin{center}
\centerline{\includegraphics[scale=0.6]{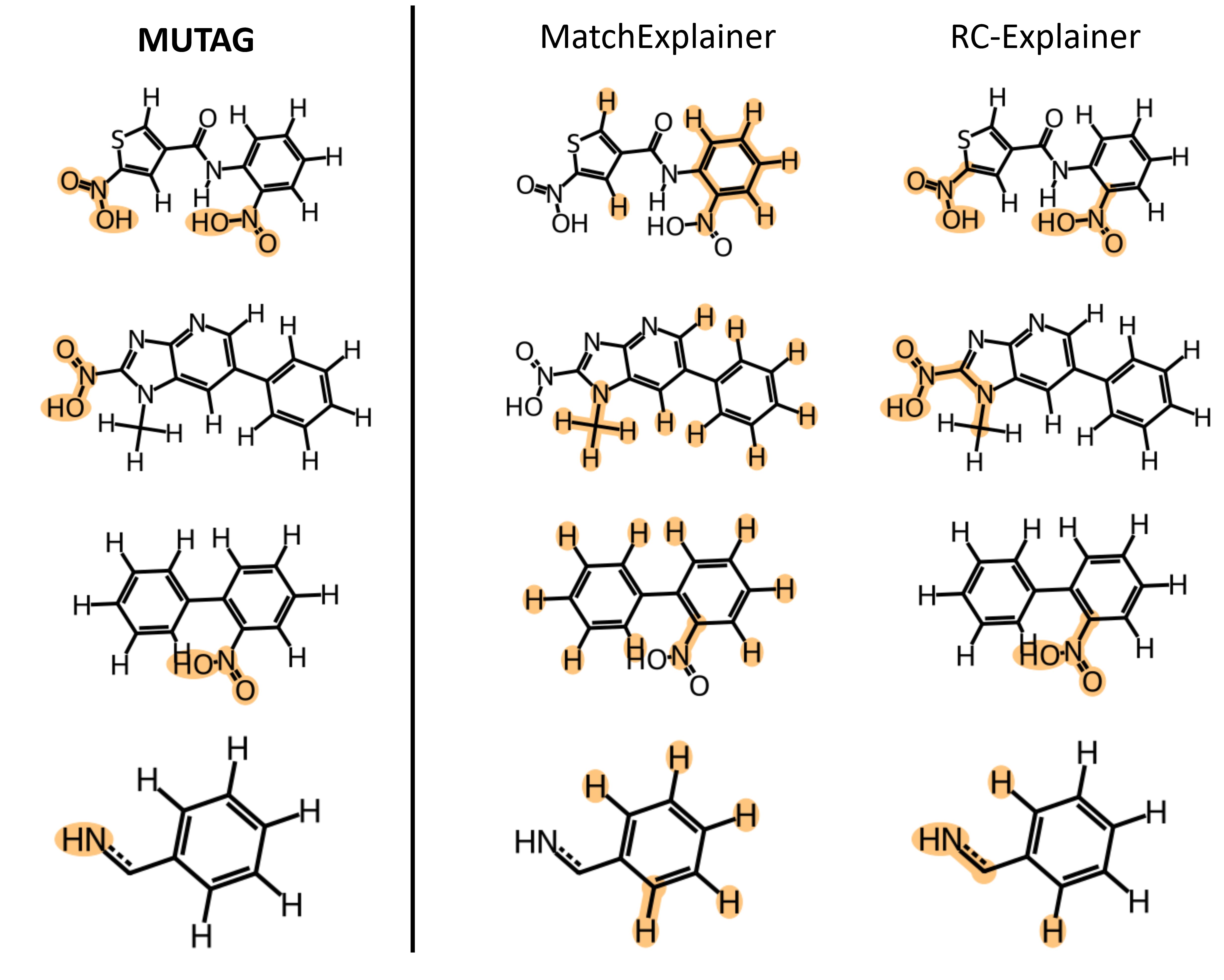}}
\caption{More examples of Explanatory subgraphs in MUTAG.}
\end{center}
\vskip -0.2in
\end{figure}

In Fig.~\ref{sup-fig-case}, we have extended the scale of qualitative analysis on MUTAG to provide a better understanding of its impact. 

\end{document}